\pdfoutput=1

\documentclass[11pt]{article}

\usepackage[preprint]{acl}

\usepackage{times}
\usepackage{latexsym}
\usepackage{graphicx} 
\usepackage{booktabs} 
\usepackage{array}    
\usepackage{caption}  
\usepackage{multirow} 

\usepackage{amsfonts}  
\usepackage{tabularx}
\usepackage{microtype}
\usepackage{graphicx}
\usepackage{subfigure}
\usepackage{booktabs} 
\usepackage{upgreek}
\usepackage{float}
\usepackage{xcolor}      
\usepackage{colortbl}
\usepackage{amsmath}
\usepackage{subcaption}
\usepackage{adjustbox}
\usepackage[T1]{fontenc}

\usepackage[utf8]{inputenc}

\usepackage{microtype}
\usepackage{tcolorbox}
\usepackage{inconsolata}

\usepackage{amsmath,amsfonts,amssymb}
\usepackage{listings}
\usepackage{textcomp}
\usepackage{colortbl}
\usepackage{xcolor}
\usepackage{tcolorbox}
\usepackage{fontawesome}

\lstdefinestyle{pytorch}{
    language=Python,
    basicstyle=\ttfamily\small,
    keywordstyle=\color{violet}\bfseries,
    commentstyle=\color{teal},
    stringstyle=\color{orange},
    numberstyle=\tiny\color{gray},
    numbers=left,                     
    stepnumber=1,
    frame=lines,                       
    breaklines=true,                    
    showstringspaces=false,
    tabsize=4
}
\definecolor{customgreen}{HTML}{006400} 
\definecolor{customblue}{HTML}{000080}  
\definecolor{customred}{HTML}{800000}   
\title{Towards Understanding and Improving Refusal in Compressed Models via Mechanistic Interpretability}

\author{Vishnu Kabir Chhabra \\
 The Ohio State University  \\
Columbus, OH \\
  \texttt{chhabra.67@osu.edu} \\\And
  Mohammad Mahdi Khalili \\
 The Ohio State University  \\
Columbus, OH\\
  \texttt{khalili.14@osu.edu} \\}

\begin{document}
\maketitle
\begin{abstract}
The rapid growth of large language models has spurred significant interest in model compression as a means to enhance their accessibility and practicality. While extensive research has explored model compression through the lens of safety, findings suggest that safety-aligned models often lose elements of trustworthiness post-compression. Simultaneously, the field of mechanistic interpretability has gained traction, with notable discoveries, such as the identification of a single direction in the residual stream mediating refusal behaviors across diverse model architectures.
In this work, we investigate the safety of compressed models by examining the mechanisms of refusal, adopting a novel interpretability-driven perspective to evaluate model safety. Furthermore, leveraging insights from our interpretability analysis, we propose a lightweight, computationally efficient method to enhance the safety of compressed models without compromising their performance or utility.
\end{abstract}

\section{Introduction}
Deployed large language models undergo safety-alignment \cite{rafailov2023direct,zhou2024beyond} to ensure trustworthiness and become more helpful and less harmless \cite{bai2022training}. Furthermore, due to the scale and size of these models, compressing large language models has been an active field of research\cite{zhu2023survey,wang2024model,yao2023comprehensive}, with considerable advances in quantization \cite{xiao2023smoothquant,lin2024awqactivationawareweightquantization,shao2023omniquant}, pruning \cite{sun2023simple, frantar2023sparsegpt, ma2023llm, kurtic2023ziplm} and low-rank factorization \cite{li2023losparse, yuan2023asvd,hsu2022language}. While research in this direction has been exciting and improved model efficiency, concerns regarding the trustworthiness and safety of compressed models remain \cite{hong2024decoding}.  

To address such concerns, recent works have analyzed such compressed models in regard to their safety and trustworthiness with the general consensus indicating that safety-aligned large language models lose some aspects of their safety after undergoing compression \cite{hong2024decoding, xu2024perplexitymultidimensionalsafetyevaluation, zhu2024safety}. This compromise in safety ranges widely between the compression techniques, with recent literature \cite{hong2024decoding} indicating that quantized models enjoy improved trustworthiness over their low-rank or pruned counterparts. To the best of our knowledge, no relevant literature analyzes the cause of this discrepancy among the techniques, hence, as one of our contributions we aim to answer why quantized models are safer than their pruned counterparts. 
\begin{figure}[h]
    \centering
    \includegraphics[width=\linewidth]{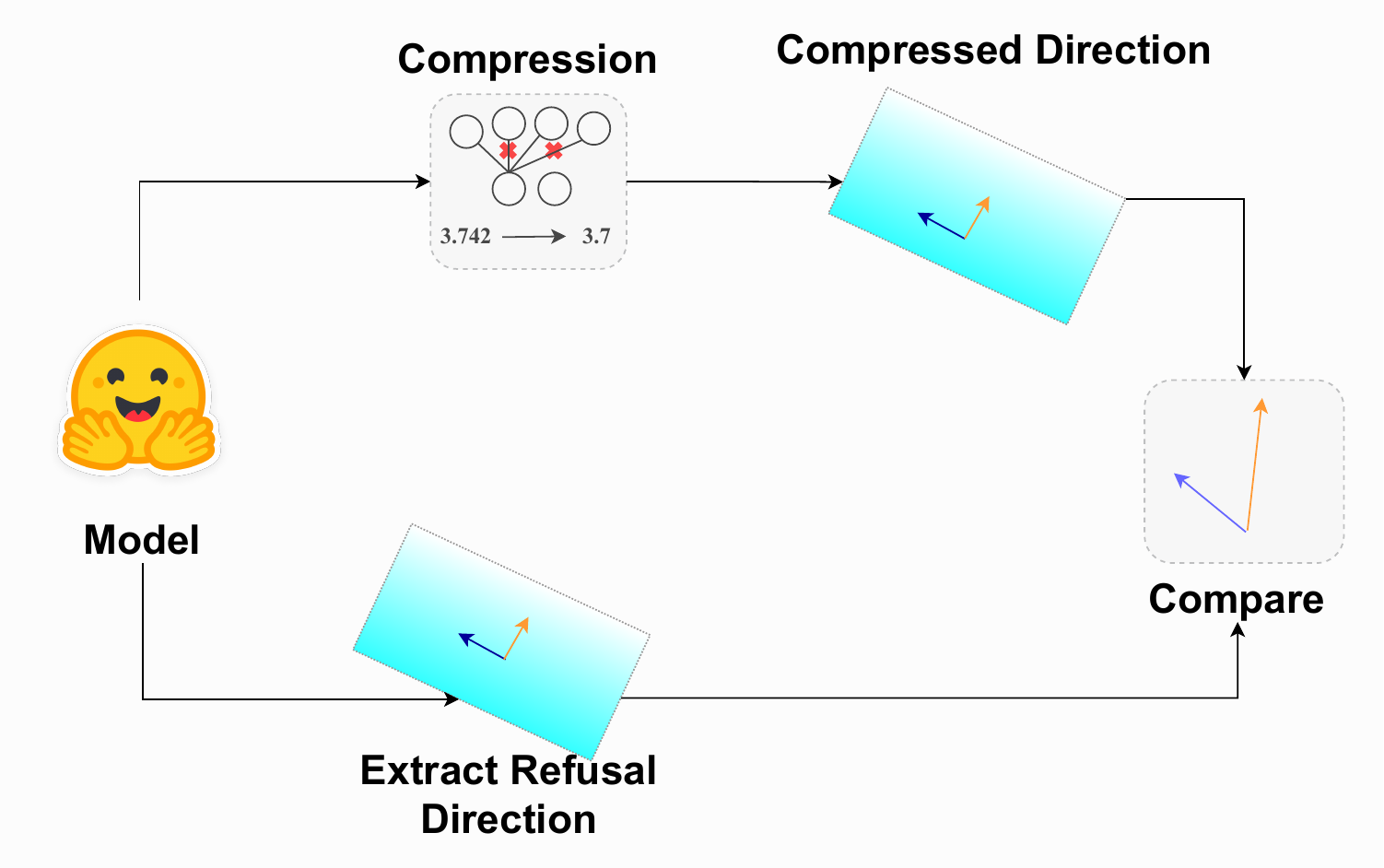}
    \caption{\textbf{Interpretability Pipeline} for comparing refusal in Compressed vs Base models.}
    \label{fig:1}
\end{figure}
Consequently, research in mechanistic interpretability has garnered attention due to the promise of decomposing the non-linear decisions of a model into human-interpretable mechanisms \cite{Olah2022,Olah2023}. Recent works have focused on reverse engineering activations into circuits \cite{wang2022interpretability, hanna2024does,merullo2023circuit,garcia2024does} that explain the functionality of the model on certain tasks, while some works have focused on understanding model decisions in niche scenarios such as grokking \cite{nanda2023progress,zhong2024clock} and some focusing on understanding the impact of fine-tuning on model mechanisms \cite{jain2024makes,prakash2024fine,chhabra2024neuroplasticity}.

\begin{figure}
    \centering
    \includegraphics[width=\linewidth]{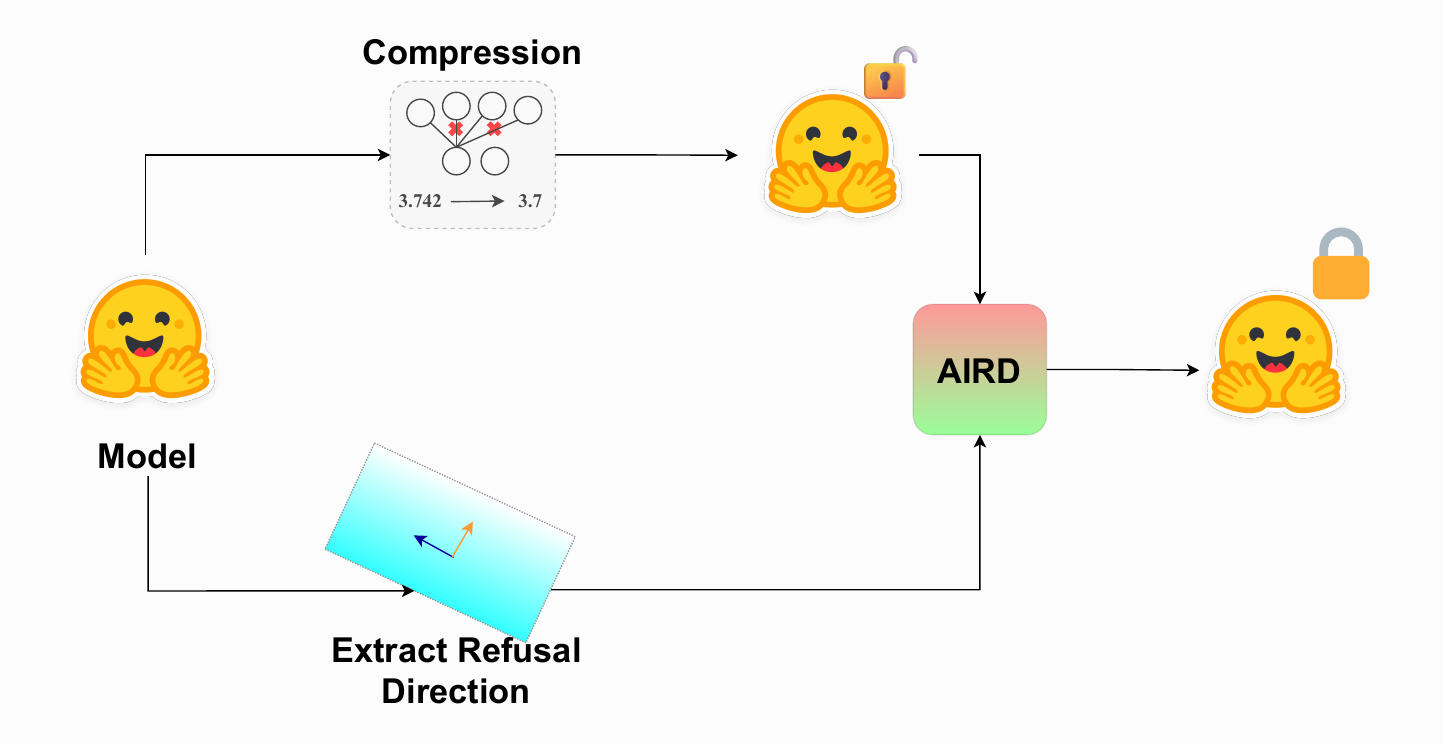}
    \caption{Artificially Inducing Refusal Direction (\textbf{AIRD}) pipeline for increasing safety of compressed models. }
    \label{fig:2}
\end{figure}

In regards to safety, work by \cite{arditi2024refusallanguagemodelsmediated} discovered that the behavior of refusal is mediated by a single direction in the residual stream activation space for modern safety-aligned large language models. Our work builds upon this work by focusing on understanding the changes to this mechanism of refusal in models compressed via a variety of methods in hopes of elucidating how the mechanisms of safety-related behavior alter after compression.  We then further investigate the importance of the mechanism of the refusal behavior and propose a novel lightweight algorithm to improve the trustworthiness of compressed models without altering their performance or utility. Our contributions can be summarized as follows: 
\begin{itemize}
    \item We investigate how the mechanism of refusal alters in compressed models. The compression methods tested belong to two categories:  pruning and quantization. \autoref{fig:1} shows our interpretability pipeline.
    \item We investigate why models compressed with quantization schemes outperform models compressed via other methods. 
    \item We utilize our findings from our investigations and propose a novel lightweight methodology for improving the trustworthiness of compressed models without any statistically significant downsides. Our method is called Artificially Inducing Refusal Direction (\textbf{AIRD}) and has been illustrated in \autoref{fig:2}.
\end{itemize}
\label{main:intro}

\section{Background}
\label{background}
\paragraph{Transformers:} Decoder-only transformers \cite{radford2019language,vaswani2017attention}  map input tokens $\mathbf{t} = (t_1, t_2, \ldots, t_n) \in \mathcal{V}^n$ to output probability distributions $\mathbf{y} = (\mathbf{y}_1, \mathbf{y}_2, \ldots, \mathbf{y}_n) \in \mathbb{R}^{n \times |\mathcal{V}|}$. Let $\mathbf{x}_i^{(l)}(\mathbf{t}) \in \mathbb{R}^{d_{\text{model}}}$ denote the residual stream activation of the token at position $i$ at the start of layer $l$. Each token's residual stream is initialized to its embedding $\mathbf{x}_i^{(1)} = \mathtt{Embed}(t_i)$, and then undergoes a series of transformations across $L$ layers. Each layer's transformation includes contributions from attention and MLP components,
\begin{align*}
\tilde{\mathbf{x}}_i^{(l)} = \mathbf{x}_i^{(l)} + \mathtt{Attn}^{(l)}(\mathbf{x}_{1:i}^{(l)}), \\
\mathbf{x}_i^{(l+1)} = \tilde{\mathbf{x}}_i^{(l)} + \mathtt{MLP}^{(l)}(\tilde{\mathbf{x}}_i^{(l)}).
\end{align*}
The final logits $\mathtt{logits}_i = \mathtt{Unembed}(\mathbf{x}_i^{(L+1)}) \in \mathbb{R}^{|\mathcal{V}|}$ are then transformed into probabilities over output tokens $\mathbf{y}_i = \mathtt{softmax}(\mathtt{logits}_i) \in \mathbb{R}^{|\mathcal{V}|}$ \cite{arditi2024refusallanguagemodelsmediated}.
\subsection{Refusal Direction}
Following \citet{arditi2024refusallanguagemodelsmediated}, we calculate the difference between the model's average activations when processing harmful versus harmless instructions to isolate the refusal direction. This technique, known as \emph{difference-in-means} \cite{belrose2023diffinmeans} isolates feature directions \cite{marks2023geometry,panickssery2023steering,tigges2023linear}.
\begin{align}\label{eq:mu}
\boldsymbol{\upmu}_i^{(l)} = \frac{1}{|\mathcal{D}_{\text{harmful}}^{\text{(train)}}|} \sum_{\mathbf{t} \in \mathcal{D}_{\text{harmful}}^{\text{(train)}}} \mathbf{x}_i^{(l)}(\mathbf{t})  \end{align}
\begin{align}\label{eq:nu}
\boldsymbol{\upnu}_i^{(l)} = \frac{1}{\lvert\mathcal{D}_{\text{harmless}}^{\text{(train)}}\rvert} \sum_{\mathbf{t} \in \mathcal{D}_{\text{harmless}}^{\text{(train)}}} \mathbf{x}_i^{(l)}(\mathbf{t}).
\end{align}

Hence, the \textit{difference-in-means} vector is as follows: $\mathbf{r}_i^{(l)} = \boldsymbol{\upmu}_i^{(l)} - \boldsymbol{\upnu}_i^{(l)}$. 
\paragraph{Selecting a single vector:}
Finding the difference-in-means vector $\mathbf{r}_{i}^{(l)}$ for each post-instruction token position $i \in I$ for $I = \{1,2,\ldots,n\}$ and layer $l \in [L]$ yields a set of $|I| \times L$ candidate vectors.
Then the most effective vector, $\mathbf{r}_{i^*}^{(l^*)}$, is chosen by evaluating
each candidate vector over validation sets $\mathcal{D}_{\text{harmful}}^{\text{(val)}}$ and $\mathcal{D}_{\text{harmless}}^{\text{(val)}}$ by measuring each candidate vector's ability to bypass refusal when ablated on $\mathcal{D}_{\text{harmful}}^{\text{(val)}}$ and to induce refusal when added on $\mathcal{D}_{\text{harmless}}^{\text{(val)}}$.
We follow the notation of \citet{arditi2024refusallanguagemodelsmediated} and denote the selected vector as $\mathbf{r}$, and its corresponding unit-norm vector as $\hat{\mathbf{r}}$.

\subsection{Model Interventions}
\paragraph{Activation addition:}
Given a difference-in-means vector \(\mathbf{r}^{(l)} \in \mathbb{R}^{d_{\text{model}}}\) derived from layer \(l\), we add the difference-in-means vector to the activations of a harmless prompt at layer $l$ and at all token positions $i \in I $. This shifts the average harmless activations towards the average harmful activations \cite{arditi2024refusallanguagemodelsmediated},
\begin{align}
\mathbf{x}^{(l)'} \leftarrow \mathbf{x}^{(l)} + \mathbf{r}^{(l)}. \label{eq:activation_addition}
\end{align}

\paragraph{Directional ablation:}
For a given direction $\hat{\mathbf{r}} \in \mathbb{R}^{d_{\text{model}}}$, we erase it from the model's representations using \emph{directional ablation} \cite{arditi2024refusallanguagemodelsmediated}.
Directional ablation suppresses the component along $\hat{\mathbf{r}}$ for every residual stream activation $\mathbf{x} \in \mathbb{R}^{d_{\text{model}}}$,
\begin{align}
    \mathbf{x}' \leftarrow \mathbf{x} - \hat{\mathbf{r}} \hat{\mathbf{r}}^{\intercal} \mathbf{x}. 
    \label{eq:projection}
\end{align} 

This operation is performed at every activation $\mathbf{x}_{i}^{(l)}$ and $\tilde{\mathbf{x}}_{i}^{(l)}$, across all layers $l$ and all token positions $i$.

\subsection{Compression}

\paragraph{Pruning:} Pruning methods of compression aim to zero out unimportant weights. A variety of methods exist that aim to utilize only the magnitude of weights \cite{han2015learning,han2015deep,frantar2023sparsegpt}, and those that consider weights and activations (e.g., \textbf{Wanda} method) \cite{sun2023simple}. Due to its efficiency, popularity, and minimal degradation of performance \cite{sun2023simple}, Wanda, is a center point of this study. In each layer, Wanda utilizes a pruning metric that assesses weight importance as follows,
\begin{align}
    S_{ij} = |{W_{ij}}|\cdot ||{X_j}||_2,
\end{align}
Where $||{X_j}||_2$ is $l_2$ norm of $j$th feature across different tokens and different inputs in a batch, and  $W_{ij}$ is an element in row $i$ and column $j$ of weight matrix $W$. 

\paragraph{Quantization:} Quantization is a form of compression that relies on lowering the precision of the model weights to compress the model \cite{zhu2023survey}. Modern literature primarily contains two forms of quantization: Training Aware Quantization \cite{chen2024efficientqat}, and Post-Training Quantization \cite{yao2023comprehensive}. The scope of this study contains models compressed via Post-Training Quantization techniques as Training Aware Quantization often requires fine-tuning which can lead to unintended consequences for safety \cite{qi2023fine}. Research in Post-Training Quantization has resulted in two forms of quantization methods: methods that rely on activations to assess weight importance \cite{lin2024awqactivationawareweightquantization} and weight-only quantization \cite{dettmers2022llmint88bitmatrixmultiplication}. In this work, we consider both types of quantization schemes.

\section{Experimental Setup}\label{sec:exsetup}

\paragraph{Models:} 
This study focuses on widely used safety-aligned large language models (LLMs) \cite{touvron2023llama-2, llama3modelcard} . The selected models, their parameters, and base precision for inference are listed in \autoref{tab:models}. 

\begin{table}[h]
    \centering
    \renewcommand{\arraystretch}{1.2}
    \begin{adjustbox}{max width=\linewidth}
    \begin{tabular}{llll}
        \toprule
        \textbf{Model family} & \textbf{Sizes} & \textbf{Precision} & \textbf{Reference} \\
        \midrule
        \textsc{Llama-2 Chat}       & 7B    & 16bit & \citet{touvron2023llama-2} \\
        \textsc{Llama-3 Instruct}   & 8B          & 16bit & \citet{llama3modelcard} \\
        \bottomrule
    \end{tabular}
    \end{adjustbox}
    \caption{Comparison of different model families.}
    \label{tab:models}
\end{table}

\paragraph{Compression Methods:} We examine the impact of pruning and quantization, two common compression methods, on model safety \cite{kuzmin2024pruningvsquantizationbetter}. Our pruning experiments utilize \textbf{Wanda} \cite{sun2023simple}, a popular and lightweight method that can prune language models in one-shot and does not suffer from severe performance degradation after pruning \cite{sun2023simple} and Magnitude pruning \cite{han2015learning}, a well established pruning method. As for quantization, we utilize two popular methods: \textbf{LLM.int8()}\cite{dettmers2022llmint88bitmatrixmultiplication} and Activation Aware Quantization (\textbf{AWQ}) \cite{lin2024awqactivationawareweightquantization}

\paragraph{Calibration Data for Compression:} To assess the effect and data dependency of the refusal mechanism in \textbf{activation-aware pruning}, we use two datasets with different objectives: maximizing safety and maximizing performance/utility. Following \citet{wei2024assessingbrittlenesssafetyalignment}, for safety, we utilize the \textsc{Align} dataset \cite{wei2024assessingbrittlenesssafetyalignment}, which is compiled using harmful instructions from \textsc{AdvBench} \cite{zou2023representation}, by dividing it into \textsc{AdvBench-Eval} (100 instructions for evaluation) and \textsc{AdvBench-Attr} (420 instructions for attribution). Then, the \textsc{LLama2-7b-chat} \cite{touvron2023llama} is prompted with \textsc{AdvBench-Attr}. An instruction along with the response is kept in  \textsc{AdvBench-Attr} if the LLama2-7b-chat declines providing the answer. Otherwise, the instruction will be deleted from the dataset. After finalizing   \textsc{AdvBench-Attr}, we use it for activation-aware pruning, and  \textsc{AdvBench-Eval} will be used for evaluating the safety of the pruned model.   For maximizing performance, we utilize a version of \textsc{Alpaca} \cite{taori2023alpaca} for pruning, namely, \textsc{Alpaca-Cleaned}. In \textsc{Alpaca-Cleaned}, we  excludes safety-related prompts using sensitive phrase matching \cite{qi2023fine}. For AWQ \cite{lin2024awqactivationawareweightquantization}, we follow the original methodology and use \textsc{Pile} \cite{gao2020pile} as the small calibration dataset.

\paragraph{Measuring Performance:} Following \citet{sun2023simple}, we measure the performance of the models by measuring their zero-shot accuracy on 5 tasks from EleutherAI's LM Harness \cite{eval-harness}: HellaSwag \cite{zellers2019hellaswag}, BoolQ \cite{clark2019boolq}, RTE \cite{wang2018glue}, ARC Challenge \cite{clark2018think} and Winogrande\cite{sakaguchi2021winogrande}.

\paragraph{Measuring Safety:} We measure the safety of our compressed models by evaluating its attack success rate (ASR)\footnote{Sometimes, we refer to ASR as attack score. } in response to harmful instructions. Specifically, we prompt the model using \textsc{AdvBench-{eval}}, the first 100 prompts from \textsc{AdvBench}, and collect its responses. Following \citet{zou2023universal}, we consider an attack as successful if the model’s response lacks key patterns indicative of refusal. The ASR is then computed as the ratio of successfully attacked
prompts to the total number of prompts evaluated. Following \citet{wei2024assessingbrittlenesssafetyalignment}, our safety evaluation considers three use cases: the ASR under non-malicious conditions (ASR$_\textrm{Vanilla}$),  and the ASR under two malicious settings --  ASR$_\textrm{Adv-Decoding}$  \cite{huang2023catastrophic}, where the attacker manipulates the decoding process, and  ASR$_\textrm{Adv-Suffix}$ ~\cite{zou2023universal}, where adversarial suffixes are used. Due to the high computational cost associated with calculating adversarial suffixes, we precompute several suffixes and use the three best-performed ones in our evaluation. For ASR$_\textrm{Adv-Decoding}$, we present results with and without the \texttt{[INST]} wrapper
\begin{table}[t]
\setlength{\tabcolsep}{2pt}
\centering
\resizebox{\linewidth}{!}{
\begin{tabular}{cccc}
\toprule
          &  ASR$_\textrm{Vanilla}$     & ASR$_\textrm{Adv-Suffix}$  & ASR$_\textrm{Adv-Decoding}$         \\
\midrule
Sample Times          & $1$           & $1$ & $5$         \\
System Prompt       & \faTimes   & \faTimes   & \faTimes\\
\texttt{[INST]}, \texttt{[/INST]} wrapper         & \faTimes & \faCheckSquareO    & \faTimes,\faCheckSquareO      \\
Adversarial Suffix& \faTimes & \faCheckSquareO  & \faTimes      \\
\bottomrule
\end{tabular}
}
\caption{The differences between three types of ASR in our safety evaluation.}
\label{tab:ASR_details}
\end{table}

\paragraph{Datasets for Finding and Evaluating Refusal Directions:} Following \citet{arditi2024refusallanguagemodelsmediated}, we construct $\mathcal{D}_{\text{harmful}}$ as a collection of harmful instructions from  \textsc{AdvBench} \cite{zou2023representation}, \textsc{MalicousInstruct} \cite{huang2023catastrophic}, \textsc{TDC2023} \cite{mazeika2024harmbench}, and \textsc{HarmBench} \cite{mazeika2024harmbench}. As for $\mathcal{D}_{\text{harmless}}$, we collect a set of harmless instructions from \textsc{Alpaca} \cite{taori2023alpaca}. Each $\mathcal{D}_{\text{harmful}}$ and $\mathcal{D}_{\text{harmless}}$ includes 160 samples which will be split into train and validation splits of 128 and 32 samples, respectively. We use training samples to find the refusal direction based on \eqref{eq:mu} and \eqref{eq:nu}. Then, we will use validation samples to evaluate the refusal direction through activation addition and directional ablation. 

\paragraph{Evaluation of Refusal:} Refusal is often measured in terms of substring matching the model's output with common phrases that indicate refusal. These phrases can often be "I cannot", "I am sorry", "as a Chatbot" etc.  \cite{wei2024assessingbrittlenesssafetyalignment}. We follow the prior literature \cite{lermen2023lora,liu2023autodan,robey2023smoothllm,shah2023loft,xu2023cognitive,zou2023universal} and utilize substring matching to classify outputs as refusal (refusal-score = 1) or successful attack (refusal-score= 0). To do so, we compile common substrings for each model architecture that indicate refusal
\section{Refusal under Compression}
\label{rduc}
In this section, firstly, apply several compression methods to the safety-aligned LLMs. If a compression method is data-dependent, then we use the calibration data introduced in Section \ref{sec:exsetup}. We then analyze the refusal mechanisms of models that underwent compression. We do this by utilizing $\mathcal{D}_{\text{harmful}}$ and $\mathcal{D}_{\text{harmless}}$ and difference-in-means \cite{arditi2024refusallanguagemodelsmediated,belrose2023diffinmeans} in the compressed models to first identify whether the refusal mechanism has altered. 

Surprisingly, our first finding reveals that the \textbf{refusal mechanism is still mediated by a single direction in compressed models}. We record this finding in \autoref{tab:rfuc}\footnote{\textsc{Llama3-8b-Instruct} loses a lot of performance under magnitude pruning and hence we don't present results for it.} and note that
this finding holds true for every compression method tested, model architecture/size, and calibration dataset, see \autoref{tab:rfuc}. This indicates that compressed models, even the ones that suffer from a degradation in safety and trustworthiness, retain the original mechanism by which they refuse harmful prompts.

We now validate that the refusal directions we found are enough for mediating the refusal mechanism. We do so by utilizing two model interventions: directional ablation and activation addition. 

\begin{figure}
    \centering
    \includegraphics[width=1\linewidth]{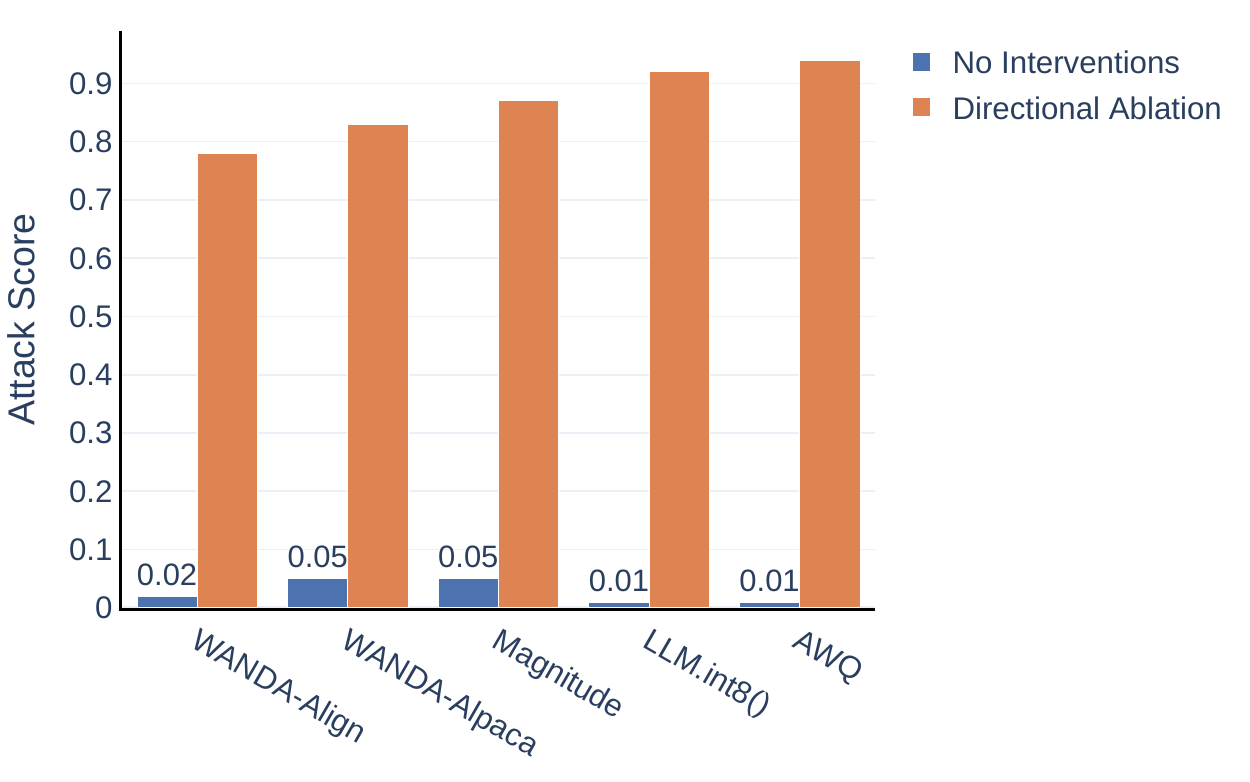}
    \caption{Attack score (\textbf{ASR}) after directional ablation in \textsc{Llama2-7b} compressed model. Ablating the refusal direction increases the attack score significantly.}
    \label{fig:dirabl}
\end{figure}

For the first model intervention, we utilize directional ablation. We borrow our methodology from the work by \citet{arditi2024refusallanguagemodelsmediated} and ablate the refusal direction from all activations at all layers and token positions. We then generate completions over $\mathcal{D}^{(\text{val})}_{\text{harmful}}$. Our findings illustrated in \autoref{fig:dirabl} show that even in the compressed models, ablating the refusal direction significantly increases the attack score of the harmful prompts. This indicates that the refusal directions we discovered for each compressed model are \textbf{necessary} for mediating refusal. Directional ablation, in this case, serves as the \textit{necessity test}, and we utilize this test to validate each refusal direction that we discover. 
\begin{table}
\centering
\resizebox{\linewidth}{!}{
 \setlength{\tabcolsep}{2pt}
\begin{tabular}{@{}llccccc@{}}
\toprule
\textbf{Type}       & \textbf{Model} & \textbf{Method} & $l^c/l$ & $i^c/i$ & \textbf{Calibration Type}  \\ 
\midrule
\multirow{4}{*}{Pruning}     & Llama2-7b        & Wanda                & $14/14$                & $\color{red}-5$$/-1$       & Alpaca        \\
                       & Llama2-7b       & Wanda              & $\color{red}12$$/14$              & $\color{red}-5$$/-1$     & Align          \\
                       & Llama2-7b        & Magnitude                &  $\color{red}12$$/14$               & $\color{red}-5$$/-1$  & ---              \\ 
            
             & Llama3-8b       & Wanda                &  $12/12$              & $-5/-5$   &Alpaca            \\
             & Llama3-8b       & Wanda                & $\color{red}13$$/12$               & $\color{red}-5$$/-1$    &Align            \\
\midrule
\multirow{3}{*}{Quantization}     & LLama2-7b        & LLM.int8()                & $14/14$               & $-1/-1$         & ---      \\
                       & LLama2-7b        & AWQ                & $14/14$              & $-1/-1$      & Pile          \\
                       & LLAma3-8b      & LLM.int8()               & $12/12$               & $-5/-5$  & ---              \\ 
                       & LLAma3-8b      & AWQ              & $12/12$                & $-5/-5$         & Pile       \\ 
\bottomrule
\end{tabular}
}
\caption{The \textbf{new refusal directions} in each compressed model tested along with the calibration dataset utilized. $l^c$, $i^c$ refer to the layer and token position of the refusal direction in the compressed model with their respective {\color{red}changes}. The \textbf{compression rate} for each method is $50\%$. }
\label{tab:rfuc}
\end{table}

\begin{figure}
    \centering
\includegraphics[width=\linewidth]{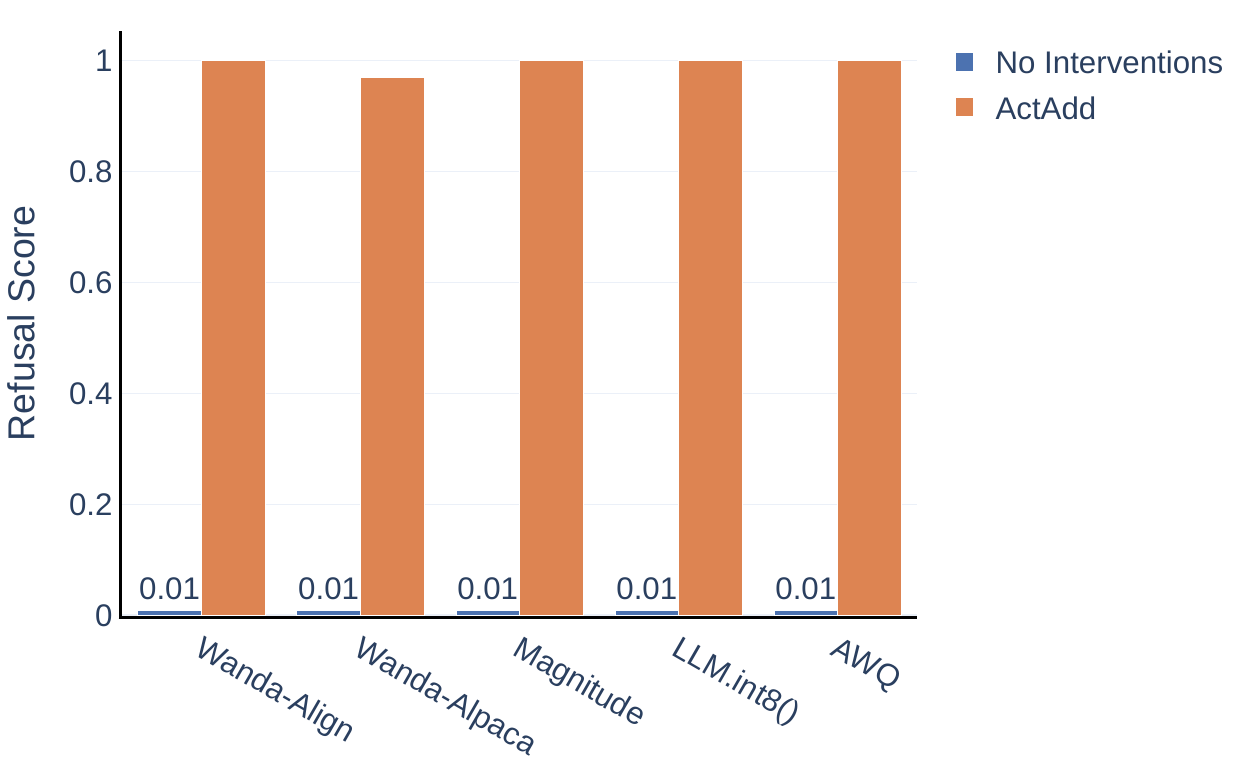}
    \caption{\textbf{Refusal Score} on harmless prompts after activation addition (\textbf{ActAdd}) in compressed \textsc{Llama2-7b} model. Activation addition causes the model to refuse to answer.}
    \label{fig:actadd}
\end{figure}

However, necessity does not imply that the refusal directions we discovered are sufficient for mediating refusal. Hence, we utilize the \textit{sufficiency test}, in which we perform activation addition at the layer $l$  and all token positions and generate completion over the validation $\mathcal{D}^{(\text{val})}_{\text{harmful}}$. 
This method, as shown by \citet{arditi2024refusallanguagemodelsmediated}, would ideally indicate that each refusal direction is sufficient for mediating refusal. Our findings (see \autoref{fig:actadd}) indeed indicate that each refusal direction we discover is \textbf{sufficient} for inducing refusal as performing activation addition on harmless instructions significantly increases the refusal score for each compression method. While \autoref{fig:dirabl} and \autoref{fig:actadd} show directional ablation and activation addition results for \textsc{Llama2-7b}, we observed similar results for \textsc{Llama3-8b}, which are omitted for brevity

Our second finding reveals that in certain compressed models, the source position of the \textbf{refusal direction changes} after compression. Surprisingly, we note that this alteration of the source position of the refusal direction is only recorded in models that are compressed via pruning methods (see \autoref{tab:rfuc}), indicating quantization of model weights has significantly less impact on a model's interpretability compared to pruning. Furthermore, for pruning, we notice an alteration in the source position of the refusal direction (for most models), indicating that the source of the refusal direction can change regardless of the calibration data.

This change in refusal direction correlates with a decrease in the trustworthiness and safety of safety-aligned models after pruning. We discuss this in the following sections.

\subsection{Does change in refusal direction mean lower safety?}
To understand the effects of the change in refusal direction in regards to safety/trustworthiness, we evaluate the performance of the compressed model on \textsc{AdvBench} \cite{zou2023representation} attacks and report our results in \autoref{tab:saf_ben}. We limit our investigation to models that undergo a change in refusal direction (either via a change in source position or the direction itself) after compression and report that the compressed models suffer from a decrease in safety across multiple dimensions of \textsc{AdvBench}\footnote{\textsc{Llama3-8b-Instruct} loses a lot of performance under magnitude pruning and hence we don't present results for it.}. This finding implies that a change in refusal direction is directly correlated with a loss of safety/trustworthiness after compression.
\begin{table}
    \centering
    \renewcommand{\arraystretch}{1.3} 
    \setlength{\tabcolsep}{1pt} 
    \resizebox{\linewidth}{!}{
    \begin{tabular}{cccccc}
        \toprule
        Model & Method  & ASR$_\textrm{Adv-Decoding}^{I}$ & ASR$_\textrm{Vanilla}$ &  ASR$_\textrm{Adv-Decoding}^{\times}$  & ASR$_\textrm{Adv-Suffix}$  \\
        \midrule
        Llama2  & Base   & $\mathbf{0.006}$  & $\mathbf{0.16}$  & $\mathbf{0.27}$ &  $\mathbf{0.09}$   \\
        Llama2  & Wanda-Align   & $\mathbf{0.0}$  & $\mathbf{0.17}$  & $\mathbf{0.26}$ &  $\mathbf{0.13}$   \\
        Llama2 & Wanda-Alpaca & $\mathbf{0.022}$ & $\mathbf{0.17}$ & $\mathbf{0.316}$ & $\mathbf{0.24}$  \\
        Llama2 & Magnitude & $\mathbf{0.01}$& $\mathbf{0.6}$ & $\mathbf{0.496}$ & $\mathbf{0.35}$\\
        Llama3  & Base     & $\mathbf{0.054}$  & $\mathbf{0.01}$  & $\mathbf{0.046}$ &  $\mathbf{0.01}$   \\
        Llama3 & Wanda-Align & $\mathbf{0.07}$& $\mathbf{0.01}$ & $\mathbf{0.076}$ & $\mathbf{0.04}$\\
        Llama3 & Wanda-Alpaca & $\mathbf{0.112}$& $\mathbf{0.04}$ & $\mathbf{0.12}$ & $\mathbf{0.13}$\\
        \bottomrule
    \end{tabular}}
    \caption{Attack Scores (\textbf{ASR}) of the compressed models on \textsc{AdvBench}.}
    \label{tab:saf_ben}
\end{table}

\section{Why Quantization is Safer than Pruning}
\citet{hong2024decoding}  noted that quantized models severely outperform their pruned counterparts across multiple dimensions of safety and trustworthiness. We further investigate this finding from an interpretability perspective. Firstly, from \autoref{tab:rfuc}, we see that quantized models do not see a change in the source of the original refusal direction whereas pruned models do see a shift in the source of the refusal direction. Secondly, as we see in Table \autoref{tab:saf_ben}, a change in the refusal direction is correlated with a loss in trustworthiness and safety. Thirdly, we measure the cosine similarity of the new refusal directions in the pruned/quantized models with the original directions. The results are provided in \autoref{tab:model_comparison} and show that the directions found in the pruned models are extremely different from the direction in the original model. The drastic shift in refusal direction observed in pruned models, but not in quantized models, explains why quantized models outperform pruned ones in safety and trustworthiness. Since quantized models retain the original refusal mechanism and its source/quality, they preserve the safety of the original model.


\begin{table}
    \centering
    \renewcommand{\arraystretch}{1.3} 
    \setlength{\tabcolsep}{6pt}      

    \rowcolors{2}{gray!15}{white}

    \begin{tabular}{l l c} 
        \toprule
        \textbf{Model} & \textbf{Method} & \textbf{Cosine Similarity} \\
        \midrule
        Llama2-7b  & Wanda-Align  & $0.351$ \\
        Llama2-7b  & LLM.int8()  & $\mathbf{0.996}$ \\
        LLama2-7b  & Wanda-Alpaca  & $0.539$ \\
        Llama2-7b  & AWQ  & $\mathbf{0.996}$ \\
        Llama2-7b  & Magnitude  & $0.337$ \\
        Llama3-8b  & Wanda-Align  & $0.732$ \\
        Llama3-8b  & LLM.int8()  & $\mathbf{0.99}$ \\
        Llama3-8b  & Wanda-Alpaca & $0.902$ \\
        LLama3-8b  & AWQ  & $\mathbf{0.994}$ \\
        \bottomrule
    \end{tabular}
    
    \caption{Comparison of different compressed models' refusal direction based on cosine similarity.}
    \label{tab:model_comparison}
\end{table}

\section{Artificially Inducing Refusal Direction}
To mitigate the effects of the altered refusal direction in the pruned models, we introduce \textbf{A}rtifically \textbf{I}nducing \textbf{R}efusal \textbf{D}irection (\textbf{AIRD}), a lightweight and simple method to increase the safety of models which suffer from an altered refusal direction after compression without loss to their performance on general coherence benchmarks. 
\paragraph{Method:} Consider a model $M$ with a refusal direction $\mathbf{r}_{i}^{(l)}$ and the compressed model $M^c$ with $\mathbf{r}_{i^{c}}^{(l^c)}$ as its  refusal direction. We orthogonalize the weight matrices that project to the residual stream (attention output and MLP output) in layer $l$  in the compressed model with respect to the refusal direction $\mathbf{r}_{i}^{(l)}$ and add it to the weight matrix as follows,
\begin{align}
    W^c_{l,new} \leftarrow W_{l}^c + \alpha \mathbf{r}_{i}^{(l)}(\mathbf{r}_{i}^{(l)})^{\intercal}W_{l}^c,
\end{align}
where $W_{l}^c$ is a weight matrix in layer $l$ of compressed model $M^c$.


\paragraph{Evaluation:} We evaluate the effects of AIRD on \textsc{AdvBench} for \textsc{LLama2-7b} and \textsc{Llama3-8b}, compressed via pruning on both calibration datasets and record our findings in \autoref{tab:aird}. 
    
\begin{table*}[!htbp]
    \centering
    \renewcommand{\arraystretch}{1.3} 
    \setlength{\tabcolsep}{2pt} 
    \begin{tabular}{ccccccc}
        \toprule
        Model & Method & Calibration  &  ASR$_\textrm{Adv-Decoding}^{I}$ & ASR$_\textrm{Vanilla}$& ASR$_\textrm{Adv-Decoding}^{\times}$  & ASR$_\textrm{Adv-Suffix}$ \\
        \midrule
        Llama2-7b  & WANDA  & Align    & {\color{customblue} $0\%$}  & {\color{customgreen} $41\%$}({\color{customgreen} $\downarrow$})  & {\color{customgreen} $14\%$}({\color{customgreen} $\downarrow$}) &  {\color{customgreen} $15\%$}({\color{customgreen} $\downarrow$})   \\
        Llama2-7b & WANDA & Alpaca  & {\color{customgreen} $10\%$}({\color{customgreen} $\downarrow$}) & {\color{customgreen} $17\%$}({\color{customgreen} $\downarrow$}) & {\color{customgreen} $12.5\%$}({\color{customgreen} $\downarrow$}) & {\color{customgreen} $41\%$}({\color{customgreen} $\downarrow$})\\
         Llama2-7b & Magnitude & ---  & {\color{customgreen} $40\%$}({\color{customgreen} $\downarrow$}) & {\color{customgreen} $20\%$}({\color{customgreen} $\downarrow$}) & {\color{customred} $2.4\%$}({\color{customred} $\uparrow$}) & {\color{customgreen} $14.2\%$}({\color{customgreen} $\downarrow$})\\
         Llama3-8b  & WANDA  & Align    & {\color{customgreen}$22.3\%$}({\color{customgreen} $\downarrow$})  & {\color{customgreen} $18.4\%$}({\color{customgreen} $\downarrow$})  & {\color{customblue} $0\%$} &  {\color{customblue} $0\%$}   \\
         Llama3-8b  & WANDA  & Alpaca   & {\color{customgreen}$10.7\%$}({\color{customgreen} $\downarrow$})  & {\color{customgreen} $33.3\%$}({\color{customgreen} $\downarrow$})  & {\color{customgreen}$17.87\%$}({\color{customgreen} $\downarrow$}) &  {\color{customgreen} $16.6\%$}({\color{customgreen} $\downarrow$})   \\
        \bottomrule
    \end{tabular}
    \caption{\textbf{Relative change} in ASR scores in models that underwent AIRD ($\downarrow$ is better). $\alpha = 0.01$ for MLP projections and $\alpha = 0.02$ for Attention projections in models.}
    \label{tab:aird}
\end{table*}
\paragraph{Core Finding:} Applying AIRD in compressed models that underwent a change in their refusal direction significantly decreases the ASR on multiple dimensions of \textsc{AdvBench}. More specifically, the attacks that succeeded more in the compressed models see a drastic change in their effectiveness against models protected with AIRD. This highlights that AIRD can successfully increase the safety of compressed models while being extremely compute efficient.

\subsection{AIRD doesn't impact performance on general benchmarks}
To understand the effect of AIRD on the general performance of the model, we evaluate the zero-shot accuracy as mentioned in \autoref{sec:exsetup}. We record our evaluations in \autoref{tab:chat_model_comparison} and find that AIRD causes no significant change in the model performance across the benchmark suite. 
Surprisingly, we find that in some benchmarks the accuracy of the compressed models increases. Although this increase is quite minuscule, it shows that our method increases the safety of the model without a significant effect on its performance. 

\begin{table*}
    \centering
    \resizebox{\textwidth}{!}{
    \renewcommand{\arraystretch}{1.5}   
    \setlength{\tabcolsep}{2pt}         

    \begin{tabular}{ l c c c c c }
        \toprule
        \textbf{Model} & \textbf{RTE} & \textbf{ARC} & \textbf{BoolQ} & \textbf{Winogrande} & \textbf{HellaSwag} \\
        \midrule
        \textbf{Llama2 Wanda-Align}   & 68.0 \color{gray}{/ 68.5} {\color{red}(-0.5)} & 36.5 \color{gray}{/ 36.0} {\color{green}(+0.5)} & 76.5 \color{gray}{/ 76} {\color{green}(+0.5)} & 64.5 \color{gray}{/ 63.0} {\color{green}(+1.5)} & 54.0 \color{gray}{/ 54.0} {\color{blue}(+0.0)} \\
        \textbf{ Llama2 Wanda-Alpaca}     & 63.5 \color{gray}{/ 64.5} {\color{red}(-1.0)} & 41.5 \color{gray}{/ 40.5} {\color{green}(+1.0)} & 79.0 \color{gray}{/ 79.0} {\color{blue}(+0.0)} & 66.0 \color{gray}{/ 66.5} {\color{red}(-0.5)} & 55.5\color{gray}{/ 55.5} {\color{blue}(+0.0)} \\
        \textbf{Llama2 Magnitude}     & 52.0 \color{gray}{/ 54.0} {\color{red}(-2.0)} & 34.5 \color{gray}{/ 34.0} {\color{red}(-0.5)} & 68.5 \color{gray}{/ 69.0} {\color{red}(-0.5)} & 61.5 \color{gray}{/ 63.0} {\color{red}(-1.5)} & 48.0\color{gray}{/ 48.0} {\color{blue}(+0.0)} \\
        \textbf{Llama3 Wanda-Align}   & 62.0 \color{gray}{/ 62.5} {\color{red}(-0.5)} & 43.5 \color{gray}{/ 44.5} {\color{red}(-0.5)} & 79.0 \color{gray}{/ 78.5} {\color{green}(+0.5)} & 70.0 \color{gray}{/ 71.0} {\color{red}(-1.0)} & 50.5 \color{gray}{/ 50.5} {\color{blue}(+0.0)} \\
        \textbf{Llama3 Wanda-Alpaca}   & 62.5 \color{gray}{/ 62.5} {\color{blue}(+0.0)} & 46.0 \color{gray}{/ 45.0} {\color{green}(+1.0)} & 82.0 \color{gray}{/ 82.0} {\color{blue}(+0.0)} & 68.5 \color{gray}{/ 67.5} {\color{green}(+1.0)} & 51.5 \color{gray}{/ 51.5} {\color{blue}(+0.0)} \\

        \bottomrule
    \end{tabular}
    }
    \caption{Performance comparison of models on the zero-shot evaluation suite. We report the zero-shot evaluations of models that underwent AIRD, {\color{gray} the base compressed model} and {\color{green}increase}, {\color{red}decrease} or {\color{blue}no change} in performance.}
    \label{tab:chat_model_comparison}
\end{table*}

\subsection{AIRD doesn't alter the refusal mechanism}
Prior work has shown that feature steering can lead to unintended consequences \cite{o2024steering,durmus2024steering}. AIRD, in this case, is a form of feature steering by orthogonalizing the weights and can possibly lead to unintentional changes in the refusal mechanism of the model. We investigate this by utilizing difference-in-means \cite{belrose2023diffinmeans} and find those models that undergo AIRD \textbf{do not see a change in the refusal mechanism}, i.e, the refusal behavior in such models is still controlled by one direction. Furthermore, we record that AIRD does not change both the source and quality of the refusal directions of models. This implies that other methods \cite{han2025internalactivationpolarstar,cao2024learnrefusemakinglarge} that rely on the present understanding of the refusal mechanism in the models are robust to the changes made by AIRD, allowing for the same degree of control as non-AIRD models

\section{Related Work}
\paragraph{Safety Under Compression:} Recent literature has explored the trustworthiness of compressed large language models via benchmarking multiple dimensions of safety \cite{hong2024decoding,xu2024perplexitymultidimensionalsafetyevaluation}, finding quantized models suffer almost no loss in safety after compression whereas pruned models do. While another work discovered the low-rank/sparse nature of safety-related components in modern LLMs \cite{wei2024assessingbrittlenesssafetyalignment}. Other works aim to improve fairness via compression \cite{xu2022can}. Although significant research progress has been made in understanding trustworthiness under compression, our work is the first of its kind, evaluating and improving compressed LLMs via mechanistic interpretability. 

\paragraph{Mechanistic Interpretability:} Through considerable manual effort, research in mechanistic interpretability has lead to important findings. Works have discovered underlying mechanisms of models as circuits \cite{wang2022interpretability,hanna2024does,merullo2023circuit,garcia2024does}, while others improve the automation of circuit discovery \cite{conmy2023towards,syed2023attribution}. Some works focus on the interpretability of mechanisms in scenarios such as grokking \cite{nanda2023progress,zhong2024clock,wang2024grokked}, fine-tuning \cite{prakash2024fine,chhabra2024neuroplasticity}. However, to the best of our knowledge, no prior work has focused on interpreting mechanisms in compressed models. 

\paragraph{Refusal Mechanism:} \citet{arditi2024refusallanguagemodelsmediated} was the first to discover that the refusal mechanism is mediated via a single direction. Follow-up works have focused on steering this refusal in large language models via feature steering \cite{o2024steering} utilizing sparse auto-encoders \cite{cunningham2023sparse}, understanding more about the refusal mechanism \cite{marshall2024refusal}, context-driven feature steering \cite{han2025internalactivationpolarstar}, introducing refusal tokens for steering \cite{jain2024refusaltokenssimpleway} and utilizing refusal for preventing hallucinations \cite{cao2024learnrefusemakinglarge}.

\section{Discussion}
In this work, we discuss the problem of understanding how a core safety-related mechanism alters in models that undergo compression. The mechanism that we discuss, the refusal mechanism, is crucial in safety against harmful prompts that seek to bypass a safety-aligned LLM's guardrails \cite{arditi2024refusallanguagemodelsmediated,o2024steering}. Our first finding indicates that in models compressed via pruning a shift in both the source position and direction occurs. However, in case of quantized models, the refusal direction retains its original characteristics. We deem this finding the source of the loss in safety that is recorded in models that are compressed via pruning. \\
Recent literature 
\cite{hong2024decoding,xu2024perplexitymultidimensionalsafetyevaluation} suggests that the quantized models don't experience any statistically significant reduction in safety after compression as opposed to their pruned counterparts, this finding resonates with our finding and we further explore this trend via comparing the directions of quantized models with that of the original uncompressed models. As directions found in quantized models retain their original characteristics and the directions in the pruned models do not, we believe this to be the mechanistic explanation as to why quantized models are safer than pruning. \\
Furthermore, based on our findings, we propose a novel lightweight, and computationally inexpensive algorithm, AIRD, that increases the safety of the compressed models that undergo a change in their refusal direction and loss in safety. Our method can increase the safety guardrails of compressed models up to $41\%$ in some benchmarks while retaining the model's coherence and not undergoing a statistically significant change in performance. Furthermore, our method doesn't impact the model's interpretability, in that both the refusal direction and the refusal mechanism are preserved in the model that undergoes AIRD. This benefit of our method implies that other techniques that rely on the refusal mechanisms\cite{cao2024learnrefusemakinglarge,o2024steering,han2025internalactivationpolarstar} can be applied to models that undergo AIRD.
\section{Limitations}
 Recent advancements in language modeling has introduced architectures that utilize Mixture of Experts \cite{jiang2023mistral, guo2025deepseek}, State Space Models \cite{gu2023mamba,lieber2024jamba}, and modernized RNNs \cite{beck2024xlstm,peng2023rwkv}. Presently, it is unclear how advancements in mechanistic interpretability for transformers and by extension our work generalize to these architectures and relevant future work is needed to generalize our findings. Furthermore, our algorithm, AIRD, reduces the compression rate of the language model by decreasing sparsity in one layer, future work can optimize and build on this work so this downside can be mitigated. 
\section{Broader Impact}
We believe mechanistic interpretability techniques
can alleviate many AI safety concerns and assist
in creating safe and reliable AI systems. However, dual-use remains a concern, as research in mechanistic interpretability can aid malicious intentions for exploiting/creating unsafe AI. However, our method, AIRD, highlights that research in the field can lead to fruitful methods that can aid in the safety of language models, but a similar method can be created to decrease a model's safety. We believe future work in this direction needs to address potential malicious side effects and create robust methods to aid in the safety and trustworthiness of language models.

\section{Acknowledgment}
This work is supported by the U.S. National Science Foundation under award
IIS-2301599 and CMMI-2301601, and by grants from the Ohio State University’s Translational Data
Analytics Institute and College of Engineering Strategic Research Initiative.
\bibliography{main}
\appendix
\newpage 
\section{Compute Statement}
All computing was performed on a cluster of 6 NVIDIA RTX A600 GPUs. The total compute time for all experiments took 200-250 hours. Reproducing experiments will take the following amount of time in a similar cluster on a single GPU:
\paragraph{Llama2-7b-chat:}
\begin{enumerate}
    \item \textit{Pruning}: Each Wanda pruning experiment takes about 10 minutes. Magnitude pruning takes 5 minutes. 
    \item \textit{Refusal Direction}: Calculating the refusal direction + evaluating via directional ablation and activation adding takes about 15 minutes.
    \item \textit{AIRD}: AIRD takes less than 10 seconds (not including model loading time).
    \item \textit{Evaluation of Zero-Shot}: Takes 10 minutes.
    \item \textit{Evaluation on AdvBench}: Takes about 15 minutes, we use vLLM \cite{kwon2023vllm} for this.
\end{enumerate}
\paragraph{Llama2-8b-chat:}
\begin{enumerate}
    \item \textit{Pruning}: Each Wanda pruning experiment takes about 10 minutes. Magnitude pruning takes 7 minutes. 
    \item \textit{Refusal Direction}: Calculating the refusal direction + evaluating via directional ablation and activation adding takes about 15 minutes.
    \item \textit{AIRD}: AIRD takes less than 10 seconds(not including model loading time).
    \item \textit{Evaluation of Zero-Shot}: Takes 10 minutes.
    \item \textit{Evaluation on AdvBench}: Takes about 15 minutes, we use vLLM \cite{kwon2023vllm} for this.
\end{enumerate}
\section{Refusal Direction Selecting Algorithm}
We borrow the refusal direction selection algorithm from \citet{arditi2024refusallanguagemodelsmediated}.
Given a collection of difference-in-means vectors, denoted as \( \{ \mathbf{r}_i^{(l)} | i \in I, l \in [L] \} \), we evaluate the following key metrics:

\begin{itemize}
    \item \textbf{\texttt{bypass\_score}}: Measures the average refusal rate on the validation set of harmful prompts (\(\mathcal{D}_{\text{harmful}}^{\text{(val)}}\)) when applying directional ablation to \( \mathbf{r}_i^{(l)} \).  
    \item \textbf{\texttt{induce\_score}}: Assesses the average refusal rate on the validation set of harmless prompts (\(\mathcal{D}_{\text{harmless}}^{\text{(val)}}\)) when the activation addition of \( \mathbf{r}_i^{(l)} \) is applied.  
    \item \textbf{\texttt{kl\_score}}: Computes the average Kullback-Leibler (KL) divergence between the model's probability distributions at the final token position when evaluated on \( \mathcal{D}_{\text{harmless}}^{\text{(val)}} \) with and without directional ablation of \( \mathbf{r}_i^{(l)} \).  
\end{itemize}

To identify the optimal direction \( \mathbf{r}_{i^*}^{(l^*)} \), we select the vector with the lowest \texttt{bypass\_score}, while ensuring the following constraints are met:  

\begin{itemize}
    \item \(\texttt{induce\_score} > 0\)  
    \begin{itemize}
        \item Ensures that the selected direction is capable of inducing a refusal response.  
    \end{itemize}
    
    \item \(\texttt{kl\_score} < 0.1\)  
    \begin{itemize}
        \item Prevents the selection of directions that excessively alter model behavior on benign prompts.  
    \end{itemize}
    
    \item \( l < 0.8L \)  
    \begin{itemize}
        \item Restricts the selection to earlier layers, avoiding interference with unembedding representations.  
    \end{itemize}
\end{itemize}

\subsection{Chat Templates}
For each model we utilize the following chat templates to prompt , see \autoref{tab:chat_templates}.

\begin{table*}[h!]
\caption{Models and their corresponding chat templates. The user instruction is denoted as \texttt{\color{blue}\{Instruction\}\color{black}}. Post-instruction tokens, as defined in \S\ref{background}, are labeled in {\color{red}red\color{black}}.}
\label{tab:chat_templates}
\centering
\begin{tabularx}{1.1\textwidth}{l X}
\toprule
Model family & Corresponding refusal phrases \\
\midrule
\textsc{Llama-2 Chat} & {\footnotesize \texttt{"[INST] \color{blue}\{Instruction\}\color{red} [/INST] \color{black}"}} \\
\textsc{Llama-3 Instruct} & {\footnotesize \parbox{0.7\textwidth}{\texttt{"<|start\_header\_id|>user<|end\_header\_id|>\string\n\string\n}\\\texttt{\color{blue}\{Instruction\}\color{red}<|eot\_id|><|start\_header\_id|>assistant<|end\_header\_id|>\string\n\string\n\color{black}"}}} \\
\bottomrule
\end{tabularx}
\end{table*}

\section{Details of Zero-Shot Evaluations}
\begin{enumerate}
    \item \textbf{ARC-Challenge:}
    \begin{enumerate}
        \item \textbf{Downstream Task:} Science Question Answering.
        \item \textbf{Overview:} This metric gauges model performance on the ARC-Challenge portion of the AI2 Reasoning Challenge dataset. It comprises grade-school science questions that necessitate complex reasoning and an in-depth understanding of scientific principles\footnote{Further details can be found at \url{https://allenai.org/data/arc}.}.
    \end{enumerate}
    \item \textbf{HellaSWAG:}
    \begin{enumerate}
        \item \textbf{Downstream Task:} Commonsense Reasoning.
        \item \textbf{Overview:} HellaSWAG is designed to test commonsense reasoning capabilities. It presents a context followed by several multiple-choice endings, with the objective of selecting the most plausible continuation. The dataset challenges models to interpret and reason about everyday situations\footnote{Additional information is available at \url{https://huggingface.co/datasets/Rowan/hellaswag}.}.
    \end{enumerate}
    \item \textbf{WinoGrande:}
    \begin{enumerate}
        \item \textbf{Downstream Task:} Commonsense Reasoning.
        \item \textbf{Overview:}  WinoGrande is a large-scale dataset for assessing commonsense reasoning. Presented as a fill-in-the-blank task with binary choices, the aim is to select the appropriate option, demanding robust commonsense understanding while mitigating dataset-specific biases\footnote{Further information is available at \url{https://huggingface.co/datasets/winogrande}.}.
    \end{enumerate}
    \item \textbf{BoolQ:}
    \begin{enumerate}
        \item \textbf{Downstream Task:} Yes/No Question Answering.
        \item \textbf{Overview:} BoolQ is a dataset focused on yes/no questions, featuring 15,942 naturally occurring examples. Each instance comprises a question, a passage, and the corresponding answer, with optional contextual information such as the page title. The setup is akin to text-pair classification tasks found in natural language inference research\footnote{More details can be found at \url{https://github.com/google-research-datasets/boolean-questions}.}.
    \end{enumerate}
    \item \textbf{RTE (Recognizing Textual Entailment):}
    \begin{enumerate}
        \item \textbf{Downstream Task:} Textual Entailment.
        \item \textbf{Overview:} The RTE task involves deciding whether a hypothesis can be logically inferred from a given premise. The dataset consists of sentence pairs, where the goal is to classify each pair as either "entailment" (if the hypothesis logically follows from the premise) or "not entailment" (if it does not)\footnote{Additional details are available at \url{https://huggingface.co/datasets/nyu-mll/glue\#rte}.}.
    \end{enumerate}
\end{enumerate}

\section{Safety Evaluations Details}
\label{sed}
\subsection{Adversarial Suffixes}
We borrow and modify the methodology of \citet{wei2024assessingbrittlenesssafetyalignment} to generate adversarial suffixes which is:
\paragraph{Llama2-7b-chat}: Run the GCG attack \cite{zou2023universal} for $500$ iterations,
with adversarial string initiated as ``\texttt{!!!!!!!!!!!!!!!!!!!!}" and a batch size of $256$, top-$k$ as $128$, with optimization over Llama2 ~\cite{touvron2023llama-2}, with the system prompts removed, for three independent trials. We then identify the top three suffixes with the highest attack success rates on AdvBench, and use them in our evaluation. 
\paragraph{Llama3-8b-instruct:} Run the GCG attack \cite{zou2023universal} for $500$ iterations, with adversarial string initiated as ``\texttt{!!!!!!!!!!!!!!!!!!!!}" and a batch size of $256$, top-$k$ as $128$, with optimization over Llama3-8b-instruct, with the system prompts removed, for three independent trials. 

For ethical reasons, we chose not to disclose the adversarial suffixes. 

\subsection{ ASR$_\textrm{Adv-Decoding}^{I}$ in Llama3-8b-instruct}
For Llama2-7b-chat we utilize the \texttt{[INST]} wrapper around the prompt for ASR$_\textrm{Adv-Decoding}^{I}$. As Llama3-8b-instruct doesn't support the \texttt{[INST]}. We modify the prompt by wrapping it around the chat template as mentioned in \autoref{tab:chat_templates}. 
\section{Sufficiency and Neccessity for Llama3-8b-Instruct}
\label{app:suff}
Following the methodology in \autoref{rduc}. We provide the sufficiency test via activation addition and neccessity test via direction ablation for Llama3-8b-instruct. 
\begin{figure}[h]
    \centering
    \includegraphics[width=\linewidth]{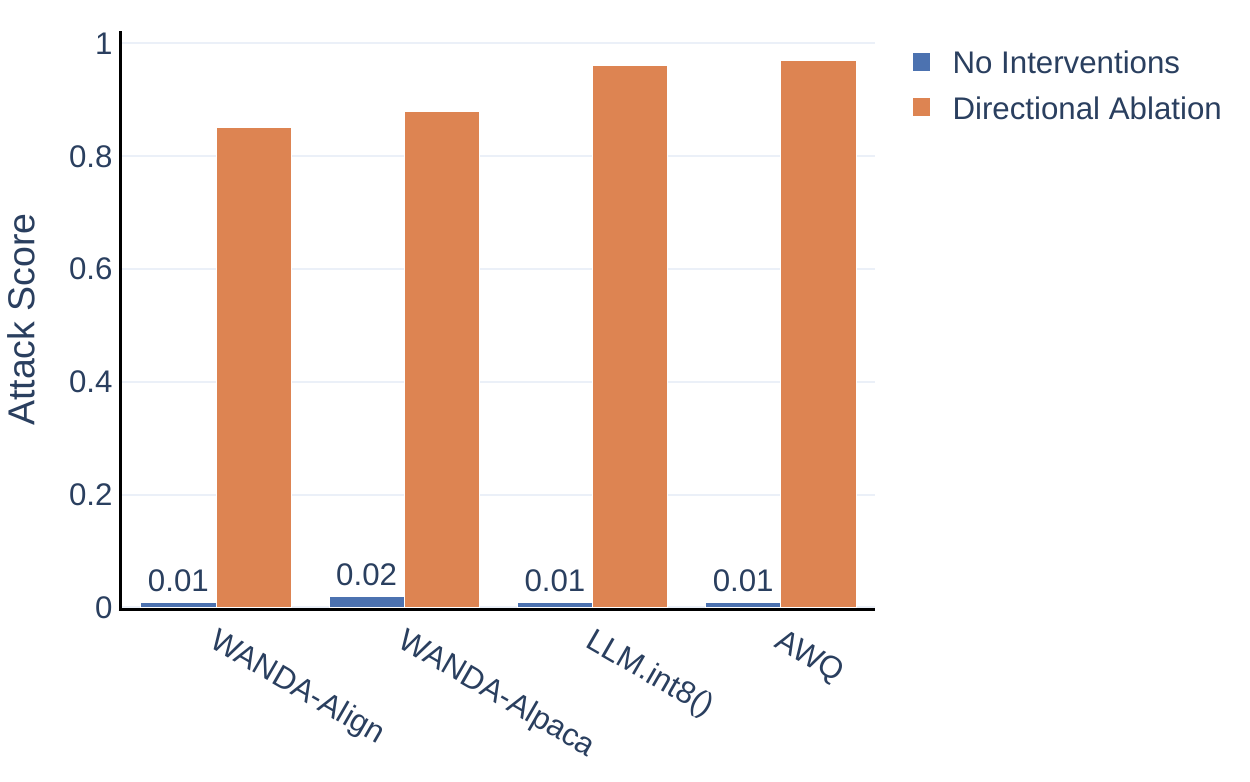}
    \caption{\textbf{Necessity Test} for Llama3-8b-instruct: Attack Score(\textbf{ASR}) after direction ablation vs no intervention on harmful instructions}
    \label{fig:app-llama3-Nec}
\end{figure}
\begin{figure}[h]
    \centering
    \includegraphics[width=\linewidth]{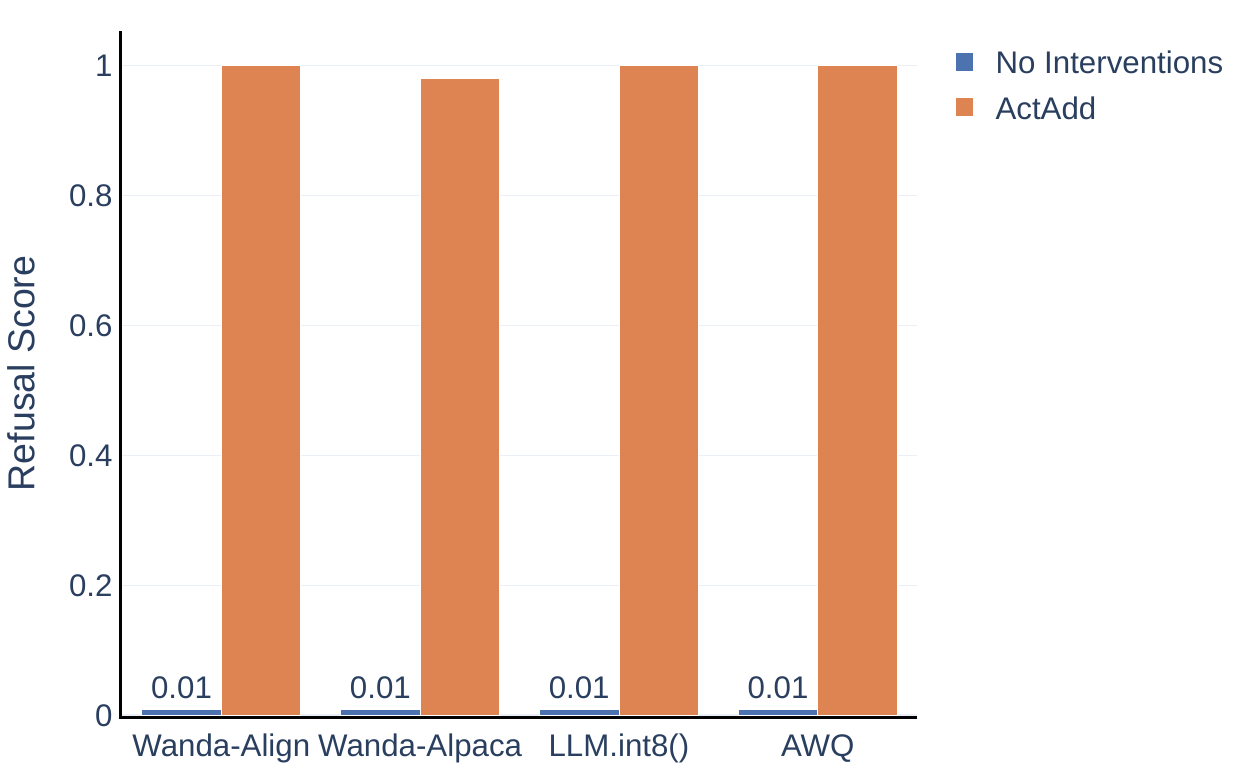}
    \caption{\textbf{Sufficiency Tes}t for Llama3-8b-instruct: Refusal Score after activation addition vs no intervention on harmless instructions}
    \label{fig:app-llama3-suff}
\end{figure}
\section{Compression Details}
\label{app:prune}
\subsection{Pruning Details}
Following the approach of \citet{sun2023simple}, we employ a block-wise pruning technique applied sequentially across Transformer blocks in Llama. Starting with the first block, we prune the seven linear layers—$\mathrm{self\_attn.q}$, $\mathrm{self\_attn.k}$, $\mathrm{self\_attn.v}$, $\mathrm{self\_attn.o}$, $\mathrm{mlp.up}$, $\mathrm{mlp.gate}$, and $\mathrm{mlp.down}$. After pruning, we recompute the block’s output before proceeding to the next one. Across all methods, we adopt unstructured pruning with a fixed sparsity of 0.5 (50\%).

\subsection{Quantization Details}
\paragraph{LLM.int8():} We utilize \url{https://huggingface.co/docs/bitsandbytes/main/en/reference/nn/linear8bit} base configuration to quantize models to 8bit. 
\paragraph{Activation Aware Quantization:} We follow the original work\footnote{\url{https://github.com/mit-han-lab/llm-awq}} to quantize models to 8bit with q\_group\_size = 128.

\section{Refusal Mechanism doesn't change after AIRD}
\label{app:rfaa}
We provide the refusal directions in the models that underwent AIRD, see \autoref{tab:pruningstuff} 
\begin{table}[h]
    \centering
    \renewcommand{\arraystretch}{1.3} 
    \setlength{\tabcolsep}{2pt} 

    \begin{tabular}{l l l c c c}
        \toprule
        Model & Method & $l^c/l$ & $i^c$/$i$ & Calibration Type \\
        \midrule
        Llama2-7b  & Wanda      & 14/14  & 5/–5  & Alpaca \\
        Llama2-7b  & Wanda      & 12/12  & 5/–5  & Align \\
        Llama2-7b  & Magnitude  & 12/12  & 5/–5  & — \\
        Llama3-8b  & Wanda      & 12/12  & 5/–5  & Alpaca \\
        Llama3-8b  & Wanda      & 13/13  & 5/–5  & Align \\
        \bottomrule
    \end{tabular}
    
    \caption{Refusal Directions of compressed models that underwent AIRD}
    \label{tab:pruningstuff}
\end{table}
As we see, no change in the directions occur. We now present the cosine similarities of the refusal directions of the model that went through AIRD with the base compressed model, see \autoref{tab:cosin}. 
\begin{table}
    \centering
    \renewcommand{\arraystretch}{1.3} 
    \setlength{\tabcolsep}{6pt}      

    \rowcolors{2}{gray!15}{white}

    \begin{tabular}{l l c} 
        \toprule
        \textbf{Model} & \textbf{Method} & \textbf{Cosine Similarity} \\
        \midrule
        Llama2-7b  & Wanda-Align  & $0.996$ \\
        LLama2-7b  & Wanda-Alpaca  & $0.995$ \\
        Llama2-7b  & Magnitude  & $0.991$ \\
        Llama3-8b  & Wanda-Align  & $0.997$ \\
        Llama3-8b  & Wanda-Alpaca & $0.997$ \\
        \bottomrule
    \end{tabular}
    
    \caption{Cosine Similarity of the refusal directions of the AIRD models and the base compressed models.}
    \label{tab:cosin}
\end{table}

\section{ASR Substrings}
\label{app:asr}
We utilize the substrings in \autoref{fig:refusal_score_substrings} for all substring matching experiments.
\begin{figure}[!h]
\begin{tcolorbox}[left=1.5mm, right=1.5mm, top=1.5mm, bottom=1.5mm]
\raggedright
{\footnotesize
\texttt{["I'm sorry", "I am sorry", "I apologize", "As an AI", "As a language model", "As an assistant", "I cannot", "I can't", "I am unable to", "I'm unable to", "I am not able to", "I'm not able to", "OpenAI", "As a Chatbot", "unethical"]}}
\end{tcolorbox}
\caption{The set of "refusal substrings" for evaluating whether a prompt is refused or not.}
\label{fig:refusal_score_substrings}
\end{figure}

\end{document}